\documentclass[letterpaper]{article} 
\usepackage{aaai23}  
\usepackage{times}  
\usepackage{helvet}  
\usepackage{courier}  
\usepackage[hyphens]{url}  
\usepackage{graphicx} 
\usepackage{natbib}  
\usepackage{caption} 
\usepackage{algorithm}
\usepackage{algorithmic}

\usepackage{amsmath}
\usepackage{amsfonts}
\usepackage{multirow}
\usepackage{booktabs}
\usepackage{makecell}

\usepackage{caption}
\usepackage{textcomp,mathcomp}
\usepackage{xcolor} 
\usepackage{arydshln}

\newcommand{\proofread}[1]{{#1}}

\usepackage{newfloat}
\usepackage{listings}
\DeclareCaptionStyle{ruled}{labelfont=normalfont,labelsep=colon,strut=off} 
\floatstyle{ruled}
\newfloat{listing}{tb}{lst}{}
\floatname{listing}{Listing}
\nocopyright

\title{\textsc{\textsc{HotCold}} Block: Fooling Thermal Infrared Detectors with\\ a Novel Wearable Design}
\author {
    Hui Wei\equalcontrib \textsuperscript{\rm 1,\rm 2},
    Zhixiang Wang\equalcontrib\textsuperscript{\rm 3,\rm 4,\rm5},
    Xuemei Jia\textsuperscript{\rm 1,\rm 2},\\
    Yinqiang Zheng\textsuperscript{\rm 3},
    Hao Tang\textsuperscript{\rm 6},
    Shin'ichi Satoh\textsuperscript{\rm 5,\rm 3},
    Zheng Wang\thanks{Corresponding Author}\textsuperscript{\rm 1,\rm 2}
}
\affiliations {
    \textsuperscript{\rm 1}National Engineering Research Center for Multimedia Software, Institute of Artificial Intelligence,\\ School of Computer Science, Wuhan University\\
    \textsuperscript{\rm 2}Hubei Key Laboratory of Multimedia and Network Communication Engineering\\
    \textsuperscript{\rm 3}The University of Tokyo \quad \textsuperscript{\rm 4}RIISE \quad \textsuperscript{\rm 5}National Institute of Informatics \quad \textsuperscript{\rm 6}CVL, ETH Zurich\\
}

\begin{document}

\maketitle

\begin{abstract}
Adversarial attacks on thermal infrared imaging expose the risk of related applications. Estimating the security of these systems is essential for safely deploying them in the real world. In many cases, realizing the attacks in the physical space requires elaborate special perturbations. These solutions are often \emph{impractical} and \emph{attention-grabbing}. To address the need for a physically practical and stealthy adversarial attack, we introduce \textsc{HotCold} Block, a novel physical attack for infrared detectors that hide persons utilizing the wearable Warming Paste and Cooling Paste. By attaching these readily available temperature-controlled materials to the body, \textsc{HotCold} Block evades human eyes efficiently. Moreover, unlike existing methods that build adversarial patches with complex texture and structure features, \textsc{HotCold} Block utilizes an SSP-oriented adversarial optimization algorithm that enables attacks with pure color blocks and explores the influence of size, shape, and position on attack performance. Extensive experimental results in both digital and physical environments demonstrate the performance of our proposed \textsc{HotCold} Block. \emph{Code is available: \textcolor{magenta}{https://github.com/weihui1308/HOTCOLDBlock}}.
\end{abstract}

\section{Introduction}
\label{sec:Introduction}

\proofread{
Deep Neural Networks (DNNs) have achieved great success in various fields. They work not only well in visible light but also in thermal infrared imaging, \emph{e.g.}, thermal infrared detection systems that are widely used in 
autonomous driving, night surveillance, temperature measurement, \emph{etc}. However, adversarial attacks in both \emph{digital} and \emph{physical} worlds expose the vulnerability of DNNs, raising concerns about the security of related applications. The security of DNNs has attracted significant attention in \emph{visible} light~\cite{xu2021towards,duan2021adversarial,cai2022context,wang2022fca,hu2022adversarial,zhong2022shadows} but has not been fully explored in thermal \emph{infrared} imaging.
}

\begin{figure}[t]
\centering
\includegraphics[width=\columnwidth]{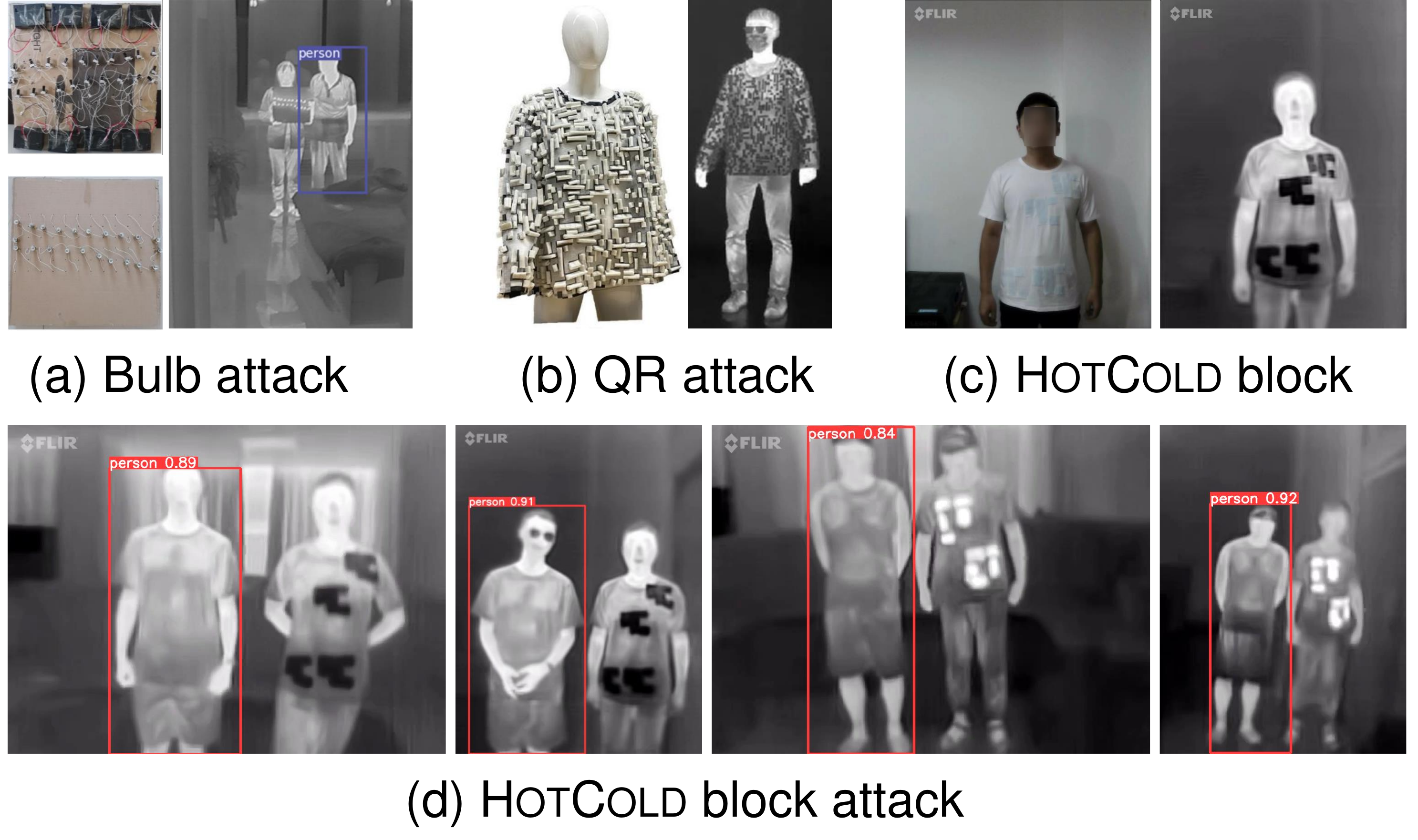} 
\caption{
\textbf{Different infrared attack methods.}
Our \textsc{HotCold} Block is effective and stealthy. It achieves competitive attack performance on YOLOv5~\cite{yolov52013} while evades human eyes better. 
}
\label{fig:teaser}
\vspace{-0.2cm}
\end{figure}

\begin{figure*}[t]
\centering
\includegraphics[width=0.95\textwidth]{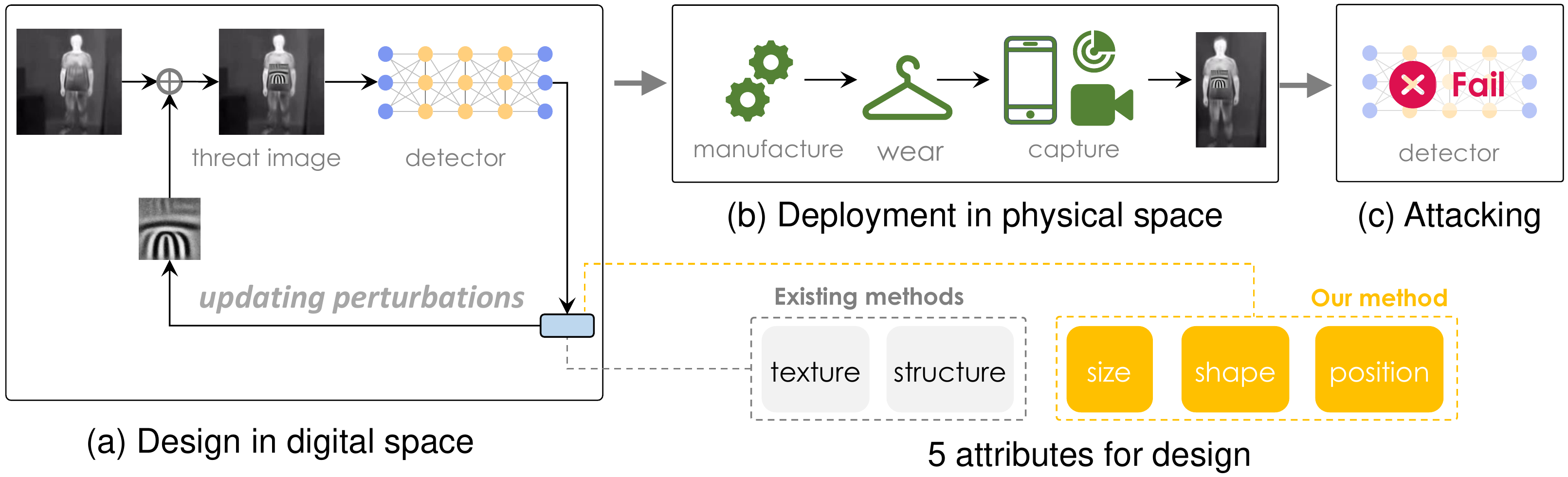} 
\caption{
\textbf{Distinguishing
the proposed method with existing methods.}
Unlike prior methods, the proposed \textsc{HotCold} Block optimizes the size, shape, and position of patches that are pervasively overlooked.}
\vspace{-0.2cm}
\label{fig:attackingStep}
\end{figure*}

\proofread{
This paper focuses on the \emph{physical} adversarial attack on \emph{infrared} detectors (for brevity, we refer to ``thermal infrared" as ``infrared" throughout the paper), which \emph{hides} persons from smart infrared cameras.
} 
Different from adversarial attack in the \emph{digital} world that directly injects adversarial perturbations to captured images, adversarial attack in the \emph{physical} world requires physically realizable objects---defined as adversarial medium---to compose adversarial perturbations. 
Recently, \citet{zhu2021fooling,zhu2022infrared} successively propose to perform attacks on infrared detectors with patches that consists of small light bulbs and invisible clothing made of aerogel. While they achieve reasonable attack effectiveness, their adversarial mediums are attention-grabbing and look unnatural to the human, making the attack 
suspicious.

To address the aforementioned problems, we 
introduce new adversarial mediums---Warming Paste and Cooling Paste---based on our findings.
First, such temperature-controlled materials can affect infrared imaging. They appear as pure color blocks under the infrared camera and interfere with detectors consequently. Second, they are wearable, making the attacks quite stealthy. Figure~\ref{fig:teaser} gives the comparison samples. Third, they are physically practical for performing the attacks, only need to paste them on the body. And finally, the Warming Paste and Cooling Paste are readily available without complicated hand-crafting. The cost to execute an attack with them is less than \$1.
Consequently, they are ideal for attacking infrared detectors, and we pose a new research question: \emph{how to design effective patterns with these new adversarial mediums}? 

This research problem is challenging since images of these materials are simple under infrared cameras and can hardly construct complex patterns. 
Existing physical adversarial attacks~\cite{thys2019fooling,liu2020bias,tan2021legitimate,hu2021naturalistic} mainly use patch-based methods, which replace a localized region of the targeted image with an elaborate patch. 
In this regard, researchers delve into the structure and texture features of the patch. They aim to generate particular patterns that can fool the DNNs. Intuitively, the attack effectiveness would suffer from the simple structure and texture of the patch. 

Considering this limitation and the imaging characteristics of the Warming Paste and Cooling Paste, we propose \textsc{HotCold} Block, which exploits pure color blocks to achieve physically practical and stealthy adversarial attacks under infrared cameras (Figure~\ref{fig:attackingStep}). To improve the performances in attack, we analyze the 5 attributes of patches: shape, size, position, structure, and texture, and design SSP-oriented adversarial optimization, which optimize the patch's {size}, {shape}, and {position} on the target human body simultaneously. To the best of our knowledge, we are the first to reveal how the lower-level attributes---size, shape, and position---compared to the high-level counterparts---texture and structure---affect the attack performance. 
Extensive experimental results on the digital and physical world demonstrate that \textsc{HotCold} Block can effectively attack infrared detectors while ensuring quite stealthy. 

Our main contributions are summarized below:
\begin{itemize}
\item We propose a new stealthy adversarial attack via the wearable temperature-sensitive materials Warming Paste and Cooling Paste, called \textsc{HotCold} Block, which is physically practical.
\item We develop an SSP-oriented adversarial optimization that considers three lower-level features of patches simultaneously, \emph{i.e.}, size, shape, and position, instead of setting them manually like most prior works.
\item We evaluate our method on mainstream detectors. Extensive experiments in digital and physical space show that our \textsc{HotCold} Block achieves competitive performance on effectiveness, stealthiness, and robustness.
\end{itemize}

\section{Related Work}
\label{sec:RelatedWork}
\subsubsection{Patch-based adversarial attacks.}
The patch-based adversarial attack is defined as an attack that is able to fool DNNs with elaborate patches and has been frequently applied to physical attacks~\cite{liu2019perceptual,liu2020bias,zolfi2021translucent}. Generally, this type of method replaces a localized region of the threat image with a patch, regardless of perturbation constraint~\cite{brown2017adversarial, liu2019dpatch, zhu2021fooling}. In recent years, researchers aimed to trade off the effectiveness and stealthiness of the patch. For instance, \citet{thys2019fooling} managed to generate smoother textures with the total variation loss. After that, \citet{wu2020making} and \citet{xu2020adversarial} printed patches on the clothing to evade human eyes, \emph{e.g.}, adversarial T-shirt and invisibility cloak. Some recent papers generated cartoon-like patches that look more natural~\cite{tan2021legitimate,hu2021naturalistic}. To sum up, previous methods mainly focus on designing special structures and textures for adversarial patches while overlooking the more general attributes: size, shape, and position. In this paper, we explore the influence of those three attributes on attack effectiveness. 

\subsubsection{Attacks to thermal infrared imaging.}
Unlike the extensive research work on adversarial attacks in visible light images~\cite{wei2022physical}, to the best of our knowledge, only two publications focus on the safety of thermal infrared imaging. \citet{zhu2021fooling} proposed a patch-based adversarial attack, which uses small glowing light bulbs to manufacture special infrared patterns. The following year, \citet{zhu2022infrared} designed infrared invisible clothing based on a new material aerogel that successfully evades person detectors. Obviously, whether small glowing light bulbs or clothes made of aerogel material, they have a common shortcoming: they are attention-grabbing when performing attacks. This is contrary to the mission of adversarial attacks. Unlike those works, we propose a physically practical and stealthy adversarial attack called \textsc{HotCold} Block. The adversarial mediums we use are the Warming Paste and Cooling Paste, wearable and readily available temperature-controlled materials.

\section{Method}
\label{sec:Method}
This section presents our method, \emph{i.e.}, \textsc{HotCold} Block.
We first introduce the modeling of the adversarial medium and then describe the SSP-oriented adversarial optimization. 

\subsection{Problem Definition}
Given an input image $I$ and the distribution of all original images $\mathcal{D}$, each $I{\in} \mathcal{D}$ contains one or multiple person instances. The pre-trained person detector $f: I {\to} \mathcal{Y}$ can predict labels $\hat{\mathcal{Y}}$ matching the true labels $\mathcal{Y}$ that includes position of the bounding boxes ${{\mathcal{V}}_{pos}}$, the object probability ${{\mathcal{V}}_{obj}}$ and the class score ${{\mathcal{V}}_{cls}}$:
\begin{equation}
\hat{\mathcal{Y}}:=[{{\mathcal{V}}_{pos}},{{\mathcal{V}}_{obj}},{{\mathcal{V}}_{cls}}]=f(I).
\end{equation}
Our goal is to fool the person detector so that it cannot identify the person, \emph{i.e.}, ${{\mathcal{V}}_{obj}}{=}0$. In this paper, we use a patch-based attack method, which replaces localized regions of the original image with patches. We denote the threat image as ${I}_{adv}$. The goal can be described as follows:
\begin{equation}
\arg \min {{\mathcal{V}}_{obj}}=\arg \underset{i}{\mathop{\min }}\,{{f}}({{I}_{adv}}),
\end{equation}
where $i$ is the index of the $i$-th image in $\mathcal{D}$.

\subsection{\textsc{HotCold} Block Modeling}
\label{subsec:Modeling}
\textsc{HotCold} Block aims to fool the infrared detector using the Warming Paste and Cooling Paste, essentially a patch-based adversarial attack. A patch generally has five attributes: size, shape, position, structure, and texture. In this paper, we delve into the size, shape, and position (``SSP" for short) of patches, with the two primary considerations: \emph{({i})} in the previous work, the lower-level features SSP of patches are pervasively set manually. Their influence on attacks has not been fully explored compared to the high level of features---texture, and structure. \emph{({ii})} since the Warming Paste and Cooling Paste are imaged as pure color blocks under the infrared camera with simple structure and texture, studying the SSP of color blocks would be more meaningful.

\begin{figure}
\centering
\includegraphics[width=\columnwidth]{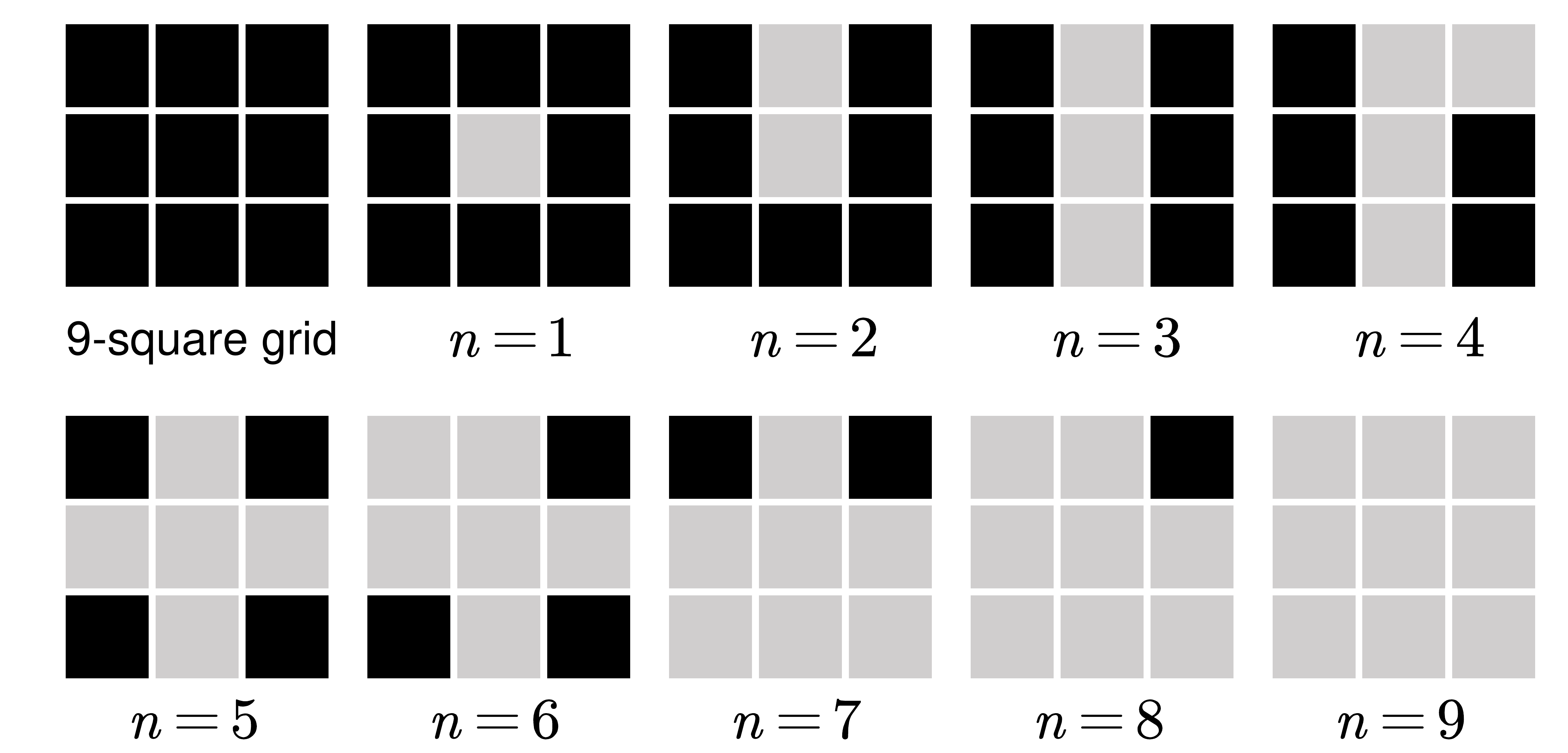} 
\caption{Example of nine-square-grid states for shape modeling. We use the optimization algorithm to find the optimal shape that minimizes the object probability ${{\mathcal{V}}_{obj}}$. As shown, the available options are diverse and flexible.}
\label{fig:shape}
\end{figure}

In practice, allowing the optimization algorithm to fit an arbitrary shape is unreasonable. One is because some shapes are not physically achievable with our adversarial medium, and the other is due to the high complexity of area calculation for some shapes. These problems are especially prominent for irregular concave polygons. Thus, we use a nine-square-grid to model our adversarial medium in the digital space, as shown in Figure \ref{fig:shape}. Consequently, not only are the aforementioned problems tackled, but we can simulate the SSP of patches reasonably and conveniently.

\begin{figure*}[t]
\centering
\includegraphics[width=0.95\textwidth]{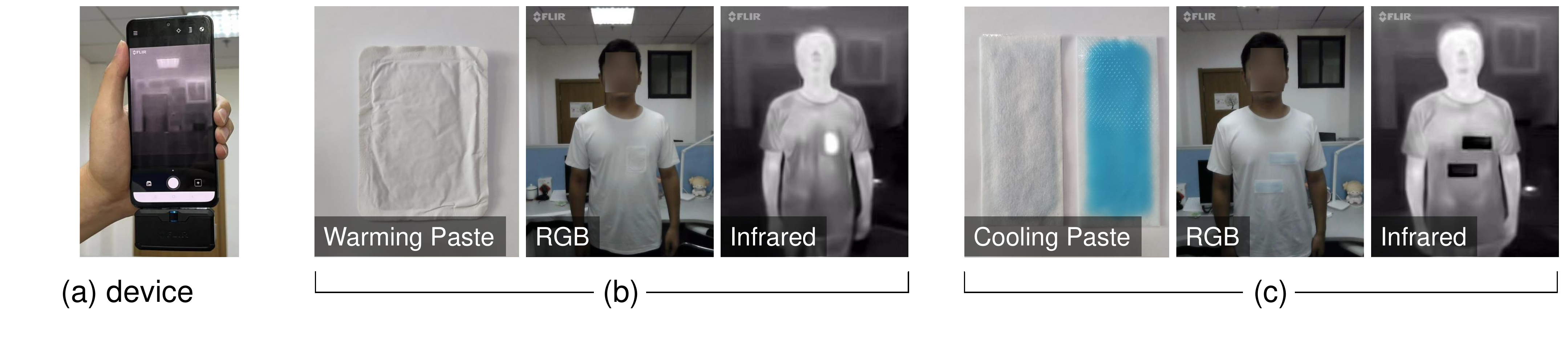} 
\caption{
\textbf{The hardware used for physical attack.}
(a) our image acquisition device. (b) the Warming Paste with their images in RGB-Infrared space. (c) the Cooling Paste with their images in RGB-Infrared space.
}
\label{fig:cameraDisplay}
\end{figure*}

\subsubsection{Size.}
Here, we define the size as the area of patches. Since \textsc{HotCold} Block uses more than one patch to attack the targeted infrared detector, the size depends on the number $m$ of patches, the number $n$ of the occupied grids, and the side length $l$ of the nine-square-grid. Note that the small size facilitates stealthiness. To minimize patch size, we design a mechanism to trade off the size and attack effectiveness. Concretely, we take the growth part of the size as a penalty term, represented as
\begin{equation}
{\mathcal{L}_{obj}}={{\mathcal{V}}_{obj}}+\underbrace{\lambda {{\Delta }_{\uparrow }}{{(\frac{n(m\times {{l}^{2}})}{9})}}}_{\text{ penalty term}}\,,
\end{equation}
where $\lambda$ is a hyperparameter to prevent ${{\mathcal{V}}_{obj}}$ from overwhelming the penalty term. The sign ${{\Delta }_{\uparrow }}$ represents positive growth, and we set the penalty term to $0$ when it decreases.

\subsubsection{Shape.}
For the shape of patches, we determine it with a $3{\times} 3$ matrix ${\mathcal{M}}$. The 0-1 value of the matrix controls the state of each grid in the nine-square-grid, \emph{e.g.}, ${\mathcal{M}}{=}[[0,1,0],[1,1,1],[0,1,0]]$ represents the state shown in Figure~\ref{fig:shape} ($n{=}5$). As shown, we can exploit flexible and complex combinations to obtain a plethora of patch shapes. The optimization algorithm in our method is dedicated to finding the optimal shape to achieve a higher attack success rate.

\subsubsection{Position.}
To improve the attack effectiveness, \textsc{HotCold} Block employs a multi-patch joint attack strategy. Our adversarial mediums, the Warming Paste and Cooling Paste, are suitable for this, and it is realizable and natural to achieve in the physical space. Here, we find an appropriate set of coordinates $\mathcal{P}{=}\{({{x}_{1}},{{y}_{1}}),({{x}_{2}},{{y}_{2}}),...,({{x}_{m}},{{y}_{m}})\}$ to determine the position of the top-left vertex of each patch, where $m$ is the number of patches. The corresponding vertices coordinate $\mathcal{P}$ is served as the parameter for optimization.

\begin{algorithm}[tb]
\caption{SSP-oriented Adversarial Optimization}
\small
\label{alg:algorithm}
\textbf{Input}: Dataset ${\mathcal{D}}$, Detector $f$.\\
\textbf{Parameter}: A vector of parameter set ${\mathcal{O}}=\{{\mathcal{M}},{\mathcal{P}}\}$\\
\textbf{Output}: ${\mathcal{M}}$, ${\mathcal{P}}$
\begin{algorithmic}[1] 
\STATE Let $t=0$.
\STATE Initialization: Randomly set ${\mathcal{M}}$, ${\mathcal{P}}$
\STATE InitializeSwarm(particles)
\FOR{$i \leftarrow 0$ to $epoch$}
\WHILE{$I=iterator({\mathcal{D}})$ is not Null}
\STATE ${{I}_{adv}}\leftarrow apply(I,{\mathcal{M}},{\mathcal{P}})$
\STATE ${\mathcal{L}_{obj}}=f({{I}_{adv}})$
\STATE ${{\mathcal{M}}^{'}},{{\mathcal{P}}^{'}}\leftarrow$ particles.$move({{\vec{x}}_{i}},{{\vec{v}}_{i}})$
\STATE $I_{adv}^{'}\leftarrow apply(I,{{\mathcal{M}}^{'}},{{\mathcal{P}}^{'}})$
\STATE ${\mathcal{L}_{obj}^{'}}=f({I_{adv}^{'}})$
\IF{${\mathcal{L}_{obj}^{'}}<{\mathcal{L}_{obj}}$}
\STATE ${\mathcal{M}}\leftarrow {{\mathcal{M}}^{'}}$, ${\mathcal{P}}\leftarrow {{\mathcal{P}}^{'}}$
\STATE particles.$update$(PersonalBest: ${{\vec{pb}}_{i}}$)
\STATE swarm.$update$(GlobalBest: ${{\vec{gb}}_{i}}$)
\ELSE
\STATE pass
\ENDIF
\ENDWHILE
\ENDFOR
\STATE \textbf{return} GlobalBest: $\{{\mathcal{M}}, {\mathcal{P}}\}$
\end{algorithmic}
\end{algorithm}

\subsection{SSP-oriented Adversarial Optimization}
In Figure~\ref{fig:cameraDisplay}, we display the adversarial mediums and their imaging under the infrared camera. Based on the aforementioned modeling, we develop an SSP-oriented adversarial optimization algorithm for performing successful attacks, with the core objectives of \textbf{{1})} minimizing patch {S}ize, \textbf{{2})} finding the optimal {S}hape, \textbf{{3})} learning suitable {P}ositions. 

Before describing the optimization algorithm, we analyze the optimization objective and optimization parameters. The focus of the optimization objective is to minimize ${\mathcal{L}_{obj}}$, aiming to result in person invisibility by detector $f$. The optimized features are the patch's size, shape, and position. Formally, we list the optimization parameter set ${\mathcal{O}}{=}\{{\mathcal{M}},{\mathcal{P}}\}$.

Since calculating a backward gradient on all operations of \textsc{HotCold} Block is challenging and the parameter values in ${\mathcal{O}}$ are discrete, solving this problem with the popular gradient descent optimization algorithm~\cite{ruder2016overview} is not appropriate. Inspired by \citet{zhong2022shadows}, we exploit the particle swarm optimization (PSO) strategy~\cite{poli2007particle}, which is a bio-inspired algorithm and does not use the gradient of the problem being optimized.

Based on the PSO, we design the SSP-oriented adversarial optimization. Specifically, a number of simple entities, the particles, are placed in the search space of our problem. Each individual in the particle swarm is composed of ${\mathcal{O}}{=}\{{\mathcal{M}},{\mathcal{P}}\}$. On each iteration, all the particles adjust their velocities ${{\vec{v}}_{i}}$ and positions ${{\vec{x}}_{i}}$. If one position is better than any that has been found so far, then the value is stored as the globe best position ${{\vec{gb}}_{i}}$ of the swarm. Meanwhile, the individual particle has its own personal best position ${{\vec{pb}}_{i}}$. The pseudocode of SSP-oriented Adversarial Optimization is shown in Algorithm~\ref{alg:algorithm}. We start with a number of random points. All the particles move in the direction of decreasing ${\mathcal{L}_{obj}}$. Each movement of particles is influenced by the ${{\vec{pb}}_{i}}$ but is also guided toward the ${{\vec{gb}}_{i}}$, which is found by the entire swarm of particles. In our optimization, the number of parameters is few, reduced by 4000 times, compared to updating the patch's structure and texture (a $300\times300$ patch that has $9\times{10}^4$ update pixels).

\section{Experiments}
\label{sec:Experiments}

In this section, we carefully evaluate the performance of our \textsc{HotCold} Block on the three criteria: \emph{effectiveness}, \emph{stealthiness}, and \emph{robustness}.

\begin{table*}[t]
\centering
\tabcolsep=6pt
\resizebox{0.88\textwidth}{!}{
\begin{tabular}{cccccccccccccccc}
\toprule
\multirow{3}{*}{\makecell[c]{Num of \\ Patches \\$m$}} &  \multirow{3}{*}{Method}  & \multicolumn{14}{c}{Side length $l$ (\%)} \\
\cmidrule(lr){3-16}
& & \multicolumn{2}{c}{6} & \multicolumn{2}{c}{8} & \multicolumn{2}{c}{10} & \multicolumn{2}{c}{12} & \multicolumn{2}{c}{14} & \multicolumn{2}{c}{16} & \multicolumn{2}{c}{Average}  \\
\cmidrule(lr){3-4}  \cmidrule(lr){5-6}  \cmidrule(lr){7-8}  \cmidrule(lr){9-10}  \cmidrule(lr){11-12}  \cmidrule(lr){13-14}  \cmidrule(lr){15-16}
& & AP$\downarrow$ & ASR$\uparrow$ & AP$\downarrow$ & ASR$\uparrow$ & AP$\downarrow$ & ASR$\uparrow$ & AP$\downarrow$ & ASR$\uparrow$ & AP$\downarrow$ & ASR$\uparrow$ & AP$\downarrow$ & ASR$\uparrow$ & AP$\downarrow$ & ASR$\uparrow$  \\
\midrule
\multirow{3}{*}{1}  & R  & 94.1 & ~~1.3 & 94.4 & ~~0.7 & 93.4 & ~~0.0 & 93.3 & ~~2.2 & 93.0 & ~~2.9 & 91.1 & ~~7.0 & 93.2 & ~~2.4\\
                    & MR & 94.7 & ~~0.0 & 94.2 & ~~4.0 & 94.3 & ~~1.8 & 93.7 & ~~3.5 & 94.3 & ~~2.2 & 93.7 & ~~2.6 & 94.2 & ~~2.4\\
                    & HCB & 93.4 & ~~1.8 & 93.2 & ~~2.9 & 90.7 & ~~3.7 & 88.0 & ~~8.4 & 85.6 & ~~9.0 & 80.9 & 14.7 & \textbf{88.6} & \textbf{~~6.8}\\ 
                    \hdashline
\specialrule{0em}{1pt}{1pt}
\multirow{3}{*}{2}  & R  & 93.4 & ~~3.7 & 92.9 & ~~3.7 & 92.6 & ~~1.7 & 91.0 & ~~5.0 & 87.4 & 12.7 & 81.8 & 18.5 & 89.9 & ~~7.6 \\
                    & MR & 93.9 & ~~2.6 & 93.2 & ~~2.8 & 92.5 & ~~2.2 & 92.2 & ~~5.3 & 90.5 & ~~6.4 & 75.0 & 24.9 & 89.6 & ~~7.4 \\
                    & HCB & 92.0 & ~~5.9 & 87.4 & ~~7.5 & 81.0 & 21.5 & 70.3 & 26.8 & 65.3 & 29.2 & 54.8 & 40.9 & \textbf{75.1} & \textbf{22.0} \\ \hdashline
\specialrule{0em}{1pt}{1pt}
\multirow{3}{*}{3}  & R  & 93.3 & ~~0.0 & 92.2 & ~~1.3 & 91.2 & ~~5.5 & 86.5 & 14.9 & 82.5 & 15.4 & 76.6 & 21.3 & 87.1 & ~~9.7 \\
                    & MR & 94.0 & ~~3.1 & 93.3 & ~~2.2 & 86.4 & 13.0 & 79.0 & 19.4 & 70.0 & 33.0 & 58.8 & 35.8 & 80.3 & 17.8\\
                    & HCB & 89.4 & ~~6.6  & 86.0 & 10.8 & 70.4 & 28.6 & 58.7 & 33.8 & 52.0 & 37.8 & 38.3 & 49.2 & \textbf{65.8} & \textbf{27.8}\\ \hdashline
\specialrule{0em}{1pt}{1pt}
\multirow{3}{*}{4}  & R  & 91.5 & ~~3.1 & 89.7 & ~~4.6 & 87.6 & 11.6 & 83.7 & 17.8 & 73.8 & 25.9 & 66.1 & 35.9 & 82.1 & 16.5\\
                    & MR & 93.5 & ~~3.5 & 92.3 & ~~4.8 & 81.8 & 19.3 & 74.7 & 23.9 & 65.8 & 37.2 & 49.0 & 38.7 & 76.2 & 21.2 \\
                    & HCB & 81.6 & 13.2 & 69.5 & 25.0 & 66.6 & 31.9 & 43.0 & 40.4 & 26.6 & 59.3 & 29.3 & 59.8 & \textbf{52.8} & \textbf{38.3}\\ \hdashline
\specialrule{0em}{1pt}{1pt}
\multirow{3}{*}{5}  & R  & 92.1 & ~~1.1 & 87.6 & 10.1 & 81.7 & 19.3 & 72.5 & 26.1 & 63.7 & 36.1 & 47.2 & 46.8 & 74.1 & 23.3 \\
                    & MR & 91.8 & ~~2.9 & 86.8 & ~~9.9 & 64.9 & 31.4 & 48.3 & 47.9 & 28.2 & 69.4 & 14.1 & 77.8 & 55.7 & 40.0 \\
                    & HCB & 83.5 & 13.6 & 58.4 & 38.3 & 49.8 & 47.2 & 34.6 & 58.7 & 22.9 & 68.8 & 11.6 & 77.8 & \textbf{43.5} & \textbf{50.7}\\ \hdashline
\specialrule{0em}{1pt}{1pt}
\multirow{3}{*}{6}  & R  & 90.7 & ~~7.5 & 87.6 & 10.1 & 78.4 & 19.1 & 64.0 & 33.0 & 62.0 & 39.3 & 31.0 & 59.9 & 69.0 & 28.2\\
                    & MR & 90.9 & ~~7.9 & 84.9 & 13.4 & 76.8 & 19.8 & 44.8 & 50.1 & 25.9 & 67.3 & ~~8.9 & 82.6 & 55.4 & 40.2\\
                    & HCB & 75.3 & 22.9 & 59.7 & 39.3 & 40.1 & 57.4 & 35.7 & 60.9 & 11.0 & 82.2 & 10.2 & 83.5 & \textbf{38.7} & \textbf{57.7} \\ \hdashline
\specialrule{0em}{1pt}{1pt}
\multirow{3}{*}{7}  & R   & 88.2 & ~~8.8  & 81.9 & 15.8 & 68.4 & 29.7 & 59.4 & 37.9 & 48.4 & 45.1 & 25.2 & 64.6 & 61.9 & 33.7\\
                    & MR  & 88.6 & 13.6 & 83.0 & 14.3 & 69.4 & 25.7 & 28.5 & 64.2 & 14.5 & 73.0 & ~~5.7 & 88.4 & 48.3 & 46.5\\
                    & HCB & 71.9 & 26.2 & 57.4 & 34.5 & 36.1 & 56.9 & 24.0 & 66.8 & 24.8 & 71.0 & ~~5.9 & 89.9 & \textbf{36.7} & \textbf{57.6}\\
\bottomrule
\end{tabular}
}
\caption{\textbf{Quantitative results on the FLIR ADAS test set at varying setups.} We report AP (\%), ASR (\%) for our adversarial attack method \textsc{HotCold} Block (HCB) \emph{vs.} the random block attack (R) and the manual-random block attack (MR), under varying numbers $m$ of patches and side lengths $l$ (\% of the person’s height).}
\label{table1}
\end{table*}

\subsection{Experimental Settings}

\subsubsection{Datasets.} 
\textbf{1)} {Digital adversarial attack.} Following~\citet{zhu2021fooling}, we evaluate the performance of our method on the Teledyne FLIR ADAS Thermal dataset ~\cite{FLIR}\footnote{Note that since the Teledyne FLIR company released an updated version in January 2022, we use v2.0 instead of v1.0. The updated dataset not only expands labels to 15 categories \emph{vs.} 5 original categories, but also the scale of the annotated image with an +83\% increase compared to the v1.0 release.}. Infrared images were acquired with a Teledyne FLIR Tau2 (13 mm f/1.0 with a 45-degree HFOV and 37-degree VFOV). The thermal camera operated in T-linear mode. We filter the original dataset for better fitting to the patch-based adversarial attack, with two conditions of \emph{({i})} the images contain ``person'' category, \emph{({ii})} the bodies of persons in the images have a height of more than 120 pixels. Finally, 1,255 images are available, of which 878 are the training set with 1,366 eligible ``person'' labels and 377 are the testing set with 598 eligible ``person'' labels. \textbf{2)} {Physical adversarial attack.} In the physical space, we capture infrared images with a FLIR ONE Pro camera~\cite{FLIRCamera}, which has a thermal resolution of $160 {\times} 120$. During capture, the camera is connected to a Xiaomi phone for real-time image display (see Figure~\ref{fig:cameraDisplay}(a)). We build lab setups that allow for shooting distances from 0 to 4 meters, record 8 videos in different scenes, and then extract one frame per second. We capture a total of 112 images and use LabelImg\footnote{LabelImg: \textcolor{magenta}{https://github.com/heartexlabs/labelImg}} to annotate them with 224 labels. 

\subsubsection{Evaluation metrics.}
We aim to hide the person from detectors. To this end, we adopt the Average Precision (AP) metric to evaluate the performance of detectors on the threat dataset. Note that lower AP indicates stronger attack effect. In addition, the Attack Success Rate (ASR) is used to evaluate the effectiveness of our attack methods, which is defined as the percentage of positive and total samples as follows:

\begin{equation}
\text{ASR}(X)=1-\frac{1}{N}\sum\limits_{i=0}^{N}{\text{sign}(labe{{l}_{i}})},
\end{equation}

\begin{equation}
\begin{aligned}
\text{sign}(labe{{l}_{i}})=\left\{ \begin{matrix}
   1, & labe{{l}_{i}}\in {{L}_{pre}}  \\
   0, & \text{otherwise.}  \\
\end{matrix} \right.
\end{aligned}
\end{equation}
where $N$ is the number of all true positive labels detected in the dataset $X$ when there is no attack, ${{L}_{pre}}$ is the set of all labels detected under attacking. The higher the ASR, the more effective the adversarial attack method is.

\proofread{
\subsubsection{Competing methods.}
We compare our method with the only 2 methods in the field of infrared attack:
\begin{itemize}
\item \textbf{Bulb Attack}~\cite{zhu2021fooling}: a physical attack method that fools infrared pedestrian detectors
using small bulbs.
\item \textbf{QR Attack}~\cite{zhu2022infrared}: a multi-angle physical attack method that designs the adversarial ``QR code" pattern for attacking infrared detectors. 
\end{itemize}
}

\proofread{
\subsubsection{Implementation details.}
We use YOLOv5~\cite{yolov52013} as our target model since it is a fast, effective, and widely-used detector. For infrared detection, we use the pre-trained weights on the MSCOCO Dataset \cite{lin2014microsoft} as initialization and fine-tune the model on the FLIR ADAS Dataset. The AP score of the fine-tuned target model achieves $94.8\%$ on the test set. We set the population size of the particle swarm to 100 and the block's pixel value to 0.2. We conduct all experiments on a device with an NVIDIA GeForce RTX 3090 GPU, and all our codes are implemented in PyTorch. Our used Warming Paste can keep on heating for 6 hours with an average temperature of 53 $\tccentigrade$, and Cooling Paste can cool down to 24 $\tccentigrade$ for 4 hours. 
}

\begin{figure*}[t]
\centering
\includegraphics[width=0.92\textwidth]{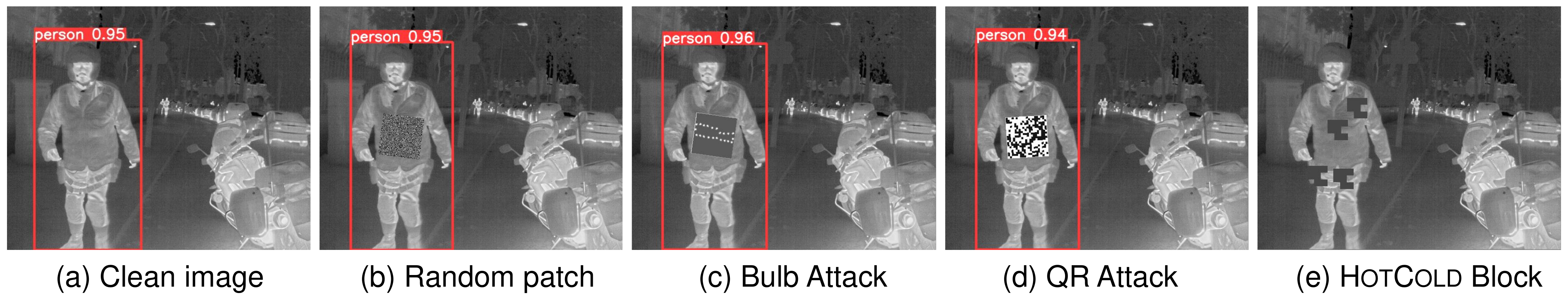} 
\caption{\textbf{Example results of digital attacks.} The bounding boxes indicate the infrared detector successfully detects the person. 
}
\label{fig:digital}
\vspace{-0.2cm}
\end{figure*}

\begin{figure}[t]
\centering
\includegraphics[width=0.95\columnwidth]{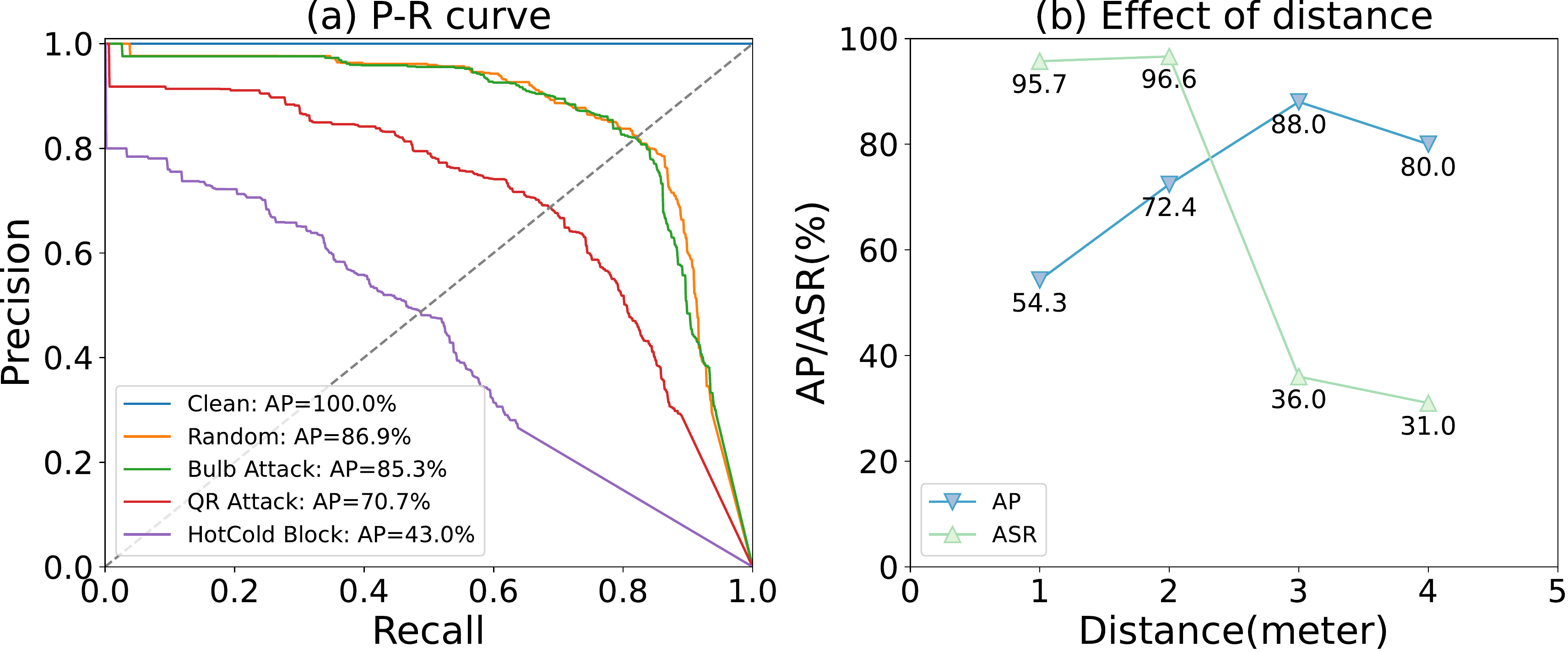} 
\caption{\textbf{Quantitative results.} (a) Precision-Recall curve 
in the digital space. (b) The AP(\%) and ASR(\%) of \textsc{HotCold} Block at different distances in the physical space.}
\label{fig:PR_curve}
\end{figure}

\subsection{Evaluation of Effectiveness}
\subsubsection{Digital adversarial attack.}
In the digital space, we attack every image in the test set of the FLIR ADAS dataset with \textsc{HotCold} Block. We run our attack with controlled numbers of patches and side lengths. Table~\ref{table1} reports the effectiveness evaluation results for our method (HCB) \emph{vs.} the random block attack (R) and the manual-random block attack (MR). MR refers to random positions, but manual intervention to avoid overlap between patches for fair comparison. Through the analysis of all experimental setups and results in Table~\ref{table1}, we can reach the following three conclusions: \textbf{{1})} \textsc{HotCold} Block comprehensively outperforms random and manual-random, demonstrating that our method is feasible and effective; \textbf{{2})} The attack effectiveness is strengthening as the number $m$ of patches and the side length $l$ increase, in keeping with our expectations; \textbf{{3})} \textsc{HotCold} Block lowers the AP to 43.0\% and achieves 40.4\% ASR under $m{=}4$ and $l{=}12$, showing that our method is physically practical due to the legitimate configuration. Note that our following experiments are in this configuration unless otherwise specified.

In Figure~\ref{fig:digital}, we show qualitative examples comparing our results with those of the baseline methods. Moreover, we draw the P-R curve for quantitative evaluation in Figure~\ref{fig:PR_curve}(a). It is clear that \textsc{HotCold} Block achieves competitive performance. For example, it causes the AP of YOLOv5 to drop by 51.8\%, significantly outperforming 9.5\% and 24.1\% achieved by the only two baseline methods. Note that, although we applied four patches for attacking (as shown in Figure~\ref{fig:digital}(e)), the total area is similar to the area of 1 patch applied in the comparison methods. 

\subsubsection{Physical adversarial attack.}
\begin{figure*}[t]
\centering
\includegraphics[width=0.8\textwidth]{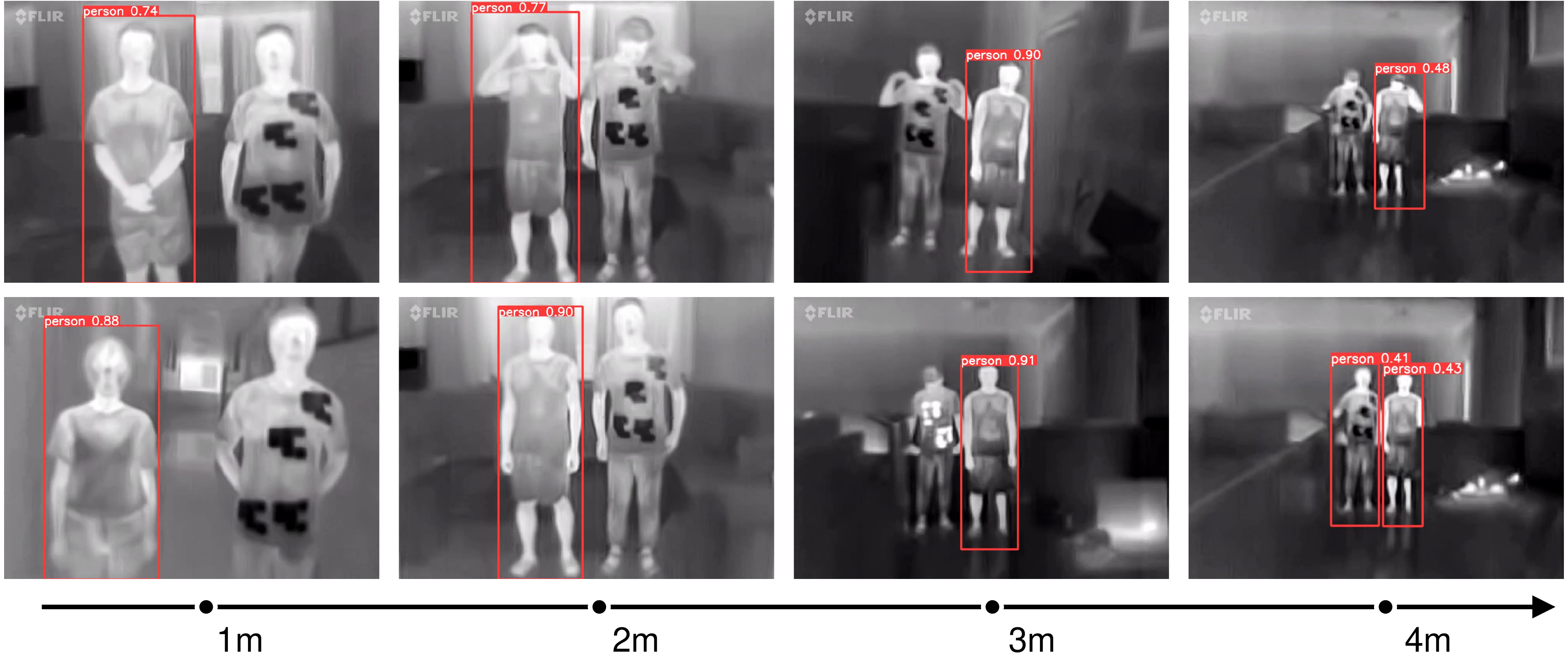} 
\caption{\textbf{Example results of physical attacks for \textsc{HotCold} Block at different distances.}}
\label{fig:physical}
\end{figure*}

Figure~\ref{fig:PR_curve}(b) and Figures~\ref{fig:physical} depict the quantitative and qualitative results, respectively. Our attack is effective and achieves an over 90\% ASR at 1m and 2m shooting distances. Although only attacking one person in the frame, the AP drops to 73.6\% on average. Observe that with the increasing shooting distance, the ASR becomes relatively low. By comparison, the shape of the patch suffers from obvious deformation due to the insufficient infrared camera resolution. This deformation may degrade the attack effectiveness. Nevertheless, the ASR is still around 34\%. The results show that \textsc{HotCold} Block is physically practical and evades infrared detectors in the real world. See \emph{Supplementary Material} for the video demo.

\subsection{Evaluation of Stealthiness}
\begin{figure}[t]
\centering
\includegraphics[width=0.94\columnwidth]{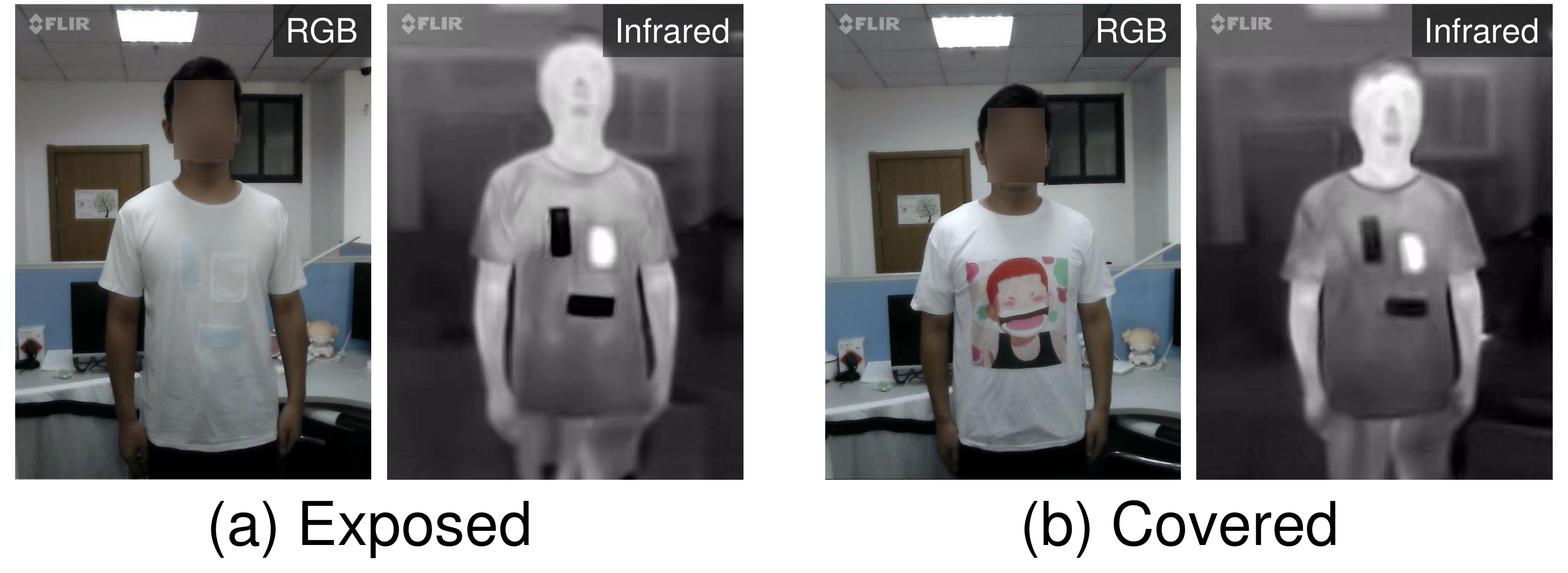} 
\caption{Examples of RGB-Infrared image pairs with exposed and covered adversarial mediums.}
\label{fig:Stealthiness}
\end{figure}

As aforementioned, considering the stealthiness of attacks, we choose the Warming Paste and Cooling Paste as our adversarial medium. In Figure~\ref{fig:Stealthiness}(a), it is clear that the adversarial mediums are in harmony with society and do not draw attention to themselves. Moreover, even if the status quo is still unsatisfactory, our method allows hiding the adversarial medium by wearing another garment on the outer surface. As shown in Figure~\ref{fig:Stealthiness}(b), the human eyes cannot recognize adversarial mediums on the human body in RGB space. Note that the change from the ``exposed'' to the ``covered'' hardly affects the infrared camera's imaging. Compared with the baseline methods, \textsc{HotCold} Block successfully achieves fairly imperceptible attacks.

\subsection{Evaluation of Robustness}
We evaluate the attack robustness of our approach across various detectors under the black box setting, including YOLOv3~\cite{redmon2018yolov3}, DETR~\cite{detr}, RetinaNet~\cite{lin2017focal}, Faster RCNN~\cite{Ren_2017}, and Mask RCNN~\cite{He_2017}. The detectors are pre-trained on the MSCOCO Dataset \cite{lin2014microsoft} and fine-tuned on the FLIR ADAS Dataset. Table~\ref{table3} reports ASR and the changes in AP. We can see that the detectors have a significant degradation in performance when facing \textsc{HotCold} Block, which makes AP drop by 44.9\% and achieves 32.2\% in ASR on average. By comparing the results, we also notice that DETR is minimally affected. One possible reason is that the transformer-based network used in DETR is beneficial in defending against adversarial attacks. Broadly, our attack is shown to be robust under the majority of models.

\begin{table}[t] \small
\centering
\resizebox{0.68\columnwidth}{!}{
\begin{tabular}{lcccc}
\toprule
\multirow{2}{*}{Detector} & \multicolumn{2}{c}{\emph{w/o} Attack} & \multicolumn{2}{c}{\emph{w/} Attack} \\
\cmidrule(lr){2-3} \cmidrule(lr){4-5}
& AP$\downarrow$ & ASR$\uparrow$ & AP$\downarrow$ & ASR$\uparrow$  \\
\midrule
YOLOv3      & 95.4 & -- & 52.5 & 40.1 \\
YOLOv5      & 94.8 & -- & 43.0 & 40.4\\
DETR        & 94.2 & -- & 65.1 & ~~4.5 \\
RetinaNet   & 93.6 & -- & 57.5 & 22.6 \\
Faster RCNN & 94.5 & -- & 31.9 & 49.4 \\
Mask RCNN   & 95.7 & -- & 48.8 & 36.3 \\
Average     & 94.7 & -- & 49.8 & 32.2 \\
\hline
\end{tabular}
}
\caption{\textbf{Evaluation across various detectors.} 
}
\label{table3}
\end{table}

\subsection{Parameter and Ablation Studies}
\subsubsection{Effect of $\lambda$.}
Larger $\lambda$ have more effect on stealthiness by reducing the size but are less effective. We evaluate the AP and ASR of our attack with different $\lambda$. As shown in Figure~\ref{fig:ablation}(a), the attack effectiveness of \textsc{HotCold} Block remains stable when $\lambda$ equals 0 to 3, and drops dramatically when $\lambda{=}4$. Moreover, We analyze the state of the nine-square-grid and observe that as $\lambda$ increases, fewer grids are activated, \emph{i.e.}, the actual area of the patch is smaller. To sum up, by modifying $\lambda$, our proposed method can trade off the visual stealthiness and effectiveness, and we set $\lambda{=}3$ to obtain the optimal balance.

\subsubsection{Effect of the block's pixel value.}
\begin{figure}[t]
\centering
\includegraphics[width=\columnwidth]{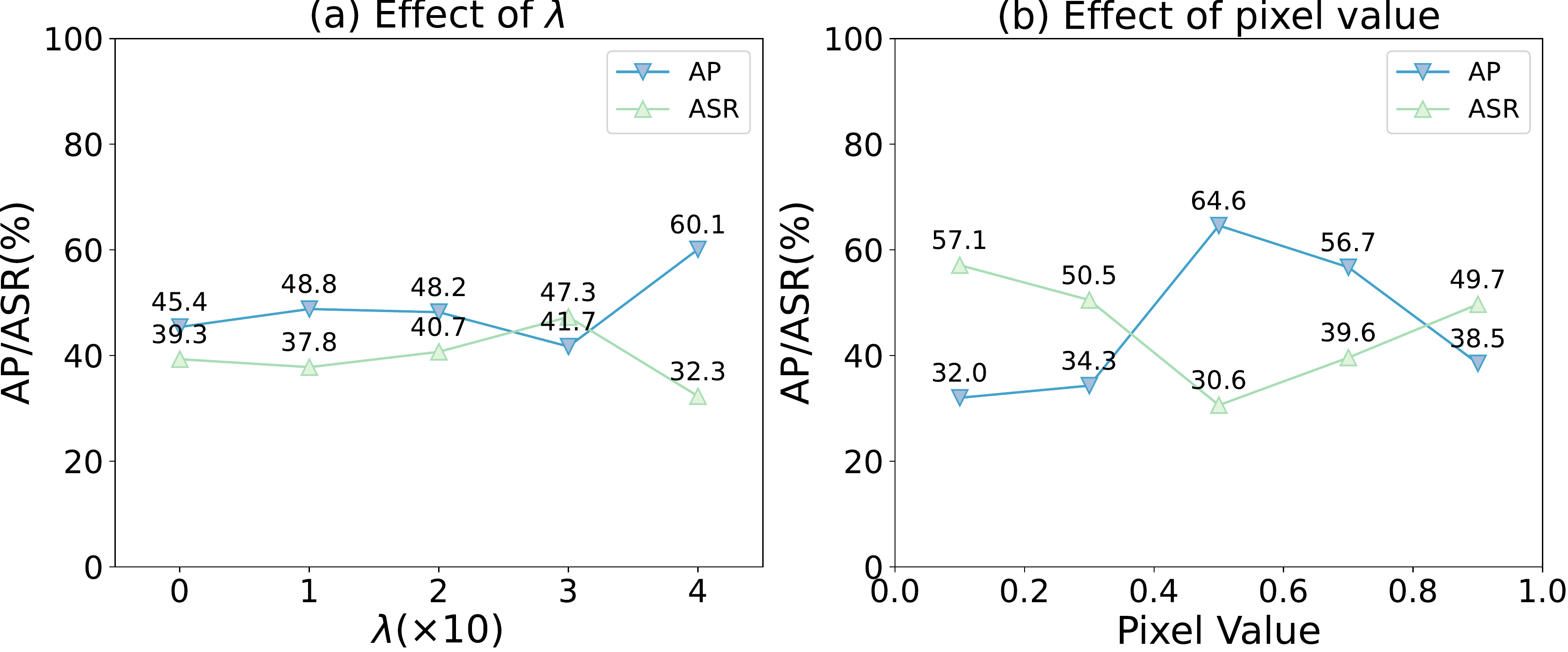} 
\caption{\textbf{Parameter and ablation studies} of hyperparameter $\lambda$ (a) and the block's pixel value (b).}
\label{fig:ablation}
\end{figure}

We then show how the block's pixel value impacts the adversarial effectiveness of \textsc{HotCold} Block. Figure~\ref{fig:ablation}(b) shows the results. Observe that as the pixel value increases, the attack capability exhibits a concave curve, which means that taking values close to 0 or 1 favors the attack. This change is because when the patch and the body are fused, \emph{i.e.}, their pixel values are close, and the patch no longer has attack capabilities. Note that the values of the Warming Paste and Cooling Paste under the infrared camera are nearly 0.9 and 0.2, respectively. Therefore, the adversarial mediums we chose are appropriate for efficient physical attacks.

\subsection{Defense Discussion}
Attack and defense develop parallel, like an arms race, to improve the model's capabilities. Here, we discuss the method for defending against \textsc{HotCold} Block. Based on our understanding of \textsc{HotCold} Block, we apply the adversarial training~\cite{ijcai2021-591} that aims to enhance the robustness of models intrinsically. Concretely, we augment training data with adversarial examples generated by \textsc{HotCold} Block in each training loop. Then we perform attacks on the retrained model to verify its effectiveness in the digital space. By comparing the results, where the AP achieves 94.9\% with no attack and 91.6\% with attacks, the retrained model becomes more robust with no performance loss. Therefore, we can use our image augmentation method to improve the detectors' performance further. In this regard, our work has important practical significance for applying DNNs-based models in the real world.

\section{Limitation}
Although \textsc{HotCold} Block shows excellent performance in attacking infrared detectors, it is hard to implement multi-angle attacks due to the \textit{segment-missing} problem~\cite{hu2022adversarial}. Concretely, since the blocks produced by the Warming Paste and Cooling Paste would be obscured when the viewing angle changes, causing them to disappear from the infrared camera, \textsc{HotCold} Block would drop in the attack success rate. We believe simulating the 3D virtual human body and sticking adversarial blocks on the surface can solve this problem.

\section{Conclusion}
In this paper, we propose a novel physical adversarial attack called \textsc{HotCold} Block that applies the Warming Paste and Cooling Paste to hide persons from being detected by infrared detectors. Our wearable adversarial mediums are physically practical and stealthy due to their intrinsic properties. Moreover, we design an SSP-oriented adversarial optimization, which delves into the feature space of size, shape, and position rather than texture and structure. Extensive experiments in both digital and physical spaces show that our \textsc{HotCold} Block evade both human eyes and detection models more effectively than existing methods.

\paragraph{Ethics statement.}
Our work successfully achieves physical attacks and illustrates the vulnerability of deep learning models. As the medium of adversarial attack is readily available and operational, most existing deep learning applications have potential security threats. Our work is dedicated to providing motivation and insights for better defense against malicious attacks.

\bibliography{aaai23}

\end{document}